\documentclass[runningheads]{llncs}
\usepackage{graphicx}
\usepackage{xcolor}
\usepackage{subfigure}
\usepackage{comment}

\begin{document}
\title{Dataset Optimization Strategies for Malware Traffic Detection}

%
\author{Ivan Letteri\inst{1}\orcidID{0000-0002-3843-386X}\inst{1} \and
Antonio Di Cecco\inst{2}\orcidID{0000-0002-9070-4663} \and
Giuseppe Della Penna\inst{1}\orcidID{0000-0003-2327-9393}}

\authorrunning{I. Letteri et al.}
%
\institute{Department of Information Engineering, Computer Science and Mathematics, University of L'Aquila, Italy \email{\{ivan.letteri,giuseppe.dellapenna\}@univaq.it} \and
School of AI, Italy \{\email{antonio.dicecco@gmail.com}\} 
}
\maketitle              
\begin{abstract}

Machine learning is rapidly becoming one of the most important technology for malware traffic detection, since the continuous evolution of malware requires a constant adaptation and the ability to generalize \cite{Letteri2018CSS}. However, network traffic datasets are usually oversized and contain redundant and irrelevant information, and this may dramatically increase the computational cost and decrease the accuracy of most classifiers, with the risk to introduce further noise. 

We propose two novel dataset optimization strategies which exploit and combine several state-of-the-art approaches in order to achieve an effective optimization of the network traffic datasets used to train malware detectors. The first approach is a feature selection technique based on mutual information measures and sensibility enhancement. The second is a dimensional reduction technique based autoencoders. Both these approaches have been experimentally applied on the  MTA-KDD'19 dataset, and the optimized results evaluated and compared using a Multi Layer Perceptron as machine learning model for malware detection.

\end{abstract}
\section{Introduction}
Machine learning (as well as any statistical methodology) faces a formidable problem when dealing with high‐dimensional data. Indeed, redundant information and variables that are not relevant for a specific classification task can dramatically increase the computational cost and decrease the accuracy of most classifiers \cite {6129431}. 
Therefore, typically the number of input variables needs to be decreased before such methodologies can be successfully applied \cite{7530147}. This operation, that we shall refer to as \textit{dataset optimization} in the rest of the paper, has three important advantages: to prevent the so called ``curse of dimensionality'' \cite{1054102}, to increase the computational efficiency of the classifiers \cite{Pasunuri2020}, also reducing the overfitting probability \cite{Gareth2013} and eases the analysis and visualization of the data \cite {Borges2012}.

This is especially true in the context of \textit{malware traffic detection}, where machine learning-powered traffic classification is rapidly becoming the only technique capable to effectively counteract the continuous evolution of malware, but the number of possible features that may be taken into account to classify the traffic is \textit{huge} and thus the datasets are often high-dimensional.

Generally speaking, dataset optimization can be accomplished in two different ways: by selecting only the most relevant variables from the original dataset (\textit{Feature Selection}, FS), or by deriving a smaller set of \textit{new} variables, as a combination of the original ones \cite{Sorzano2014}. 

Researchers have proposed many algorithms to measure the feature relevance and perform feature selection. Among them, the most promising ones seems to be based on Mutual Information (MI) \cite{LetteriPC19}, but MI can be exploited in many different ways with different results. As an example, Balagani and Phoha \cite{5432207} present an analysis of three well-known algorithms, namely mRMR \cite{1453511}, MIFS \cite{Battiti1994} and CIFE \cite{Akadi2008APF}, concluding that they make highly restrictive assumptions on the underlying data distributions. However, Brown et al. \cite{Brown2012Gavin} after comparing a number of different algorithms, suggest that JMI \cite{BENNASAR2015} and mRMR should be an optimal choice for feature selection. 

On the other hand, recently, autoencoders (AE) have shown promising results in extracting effective features from high-dimensional datasets. As an example, Shuyang et al. \cite {Wang2017Shuyang} propose a new algorithm for autoencoder-guided feature selection which tries to distinguish the task-relevant and task-irrelevant features. Kai et al.  \cite{Han2017AutoencoderFS} exploit AEs to choose the highly-representable features commonly used in a neural network for unsupervised learning. Finally,  an hybrid use of AEs applied to  feature selection is shown in \cite{Han2018}, which  combines autoencoder regression and group lasso tasks.

In this paper we define and compare two novel dataset optimization strategies which exploit and combine several of the state-of-the-art approaches described above in order to achieve an effective optimization of the network traffic datasets used to train malware detectors.

In particular, we first develop an \textit{hybrid wrapper-filter} FS strategy that tries to find the best possible subset of features with respect to a target classifier performance choosen as predictor. The naive approach to such kind of so-called \textit{wrapper} strategy is to perform an exhaustive search (i.e, experiment all the possible variables subsets), thus it tends to be computationally expensive, and often impractical when the number of variables to take into account is large. On the other hand, the proposed solution limits the subset space to explore by ranking the features through classifier-independent metrics, in particular mutual information (MI), as common in \textit{filter} strategies, to strongly limit the computation complexity.

The second optimization approach presented performs \textit{Dimensional Reduction} (DR) in order to remove inconsistent and irrelevant information from the dataset \cite{Lu2018}. DR techniques are related to FS in that both methods aim at feeding fewer input variables to the predictive model. The difference is that FS selects features to keep or remove from the dataset, whereas DR creates a projection of the data resulting in entirely new variables. 

In particular, our approach makes use of an \textit{autoencoder} (AE) \cite{HintonZemel1993} which maps (compresses) the original variables in a smaller space through nonlinear combinations, and this allows us to remove a certain amount of useless information while, at the same time, generating derived, more informative variables.
Indeed, it is well known that the learning performances of any classification algorithm are more positively influenced by the expressiveness of the features rather than by their number.



We use the MTA-KDD'19 dataset \cite{itasec2020} as a benchmark for the FS and the DR approaches, since it has a sufficiently large number of features (33) and it has already obtained excellent classification results (99.73\%). Moreover, as far as we are aware, the MTA-KDD'19 is the only public dataset aimed solely at determining malware traffic based on statistical flow analysis and which offers a constantly updated traffic collection. Most of the other public datasets are very specific to, e.g., Android or IoT malware, and sometimes based on the static analysis of executable payloads.

\section{Feature Selection through Mutual Information}
\begin{figure}[!ht]
	\centering
	\includegraphics[width=\textwidth]{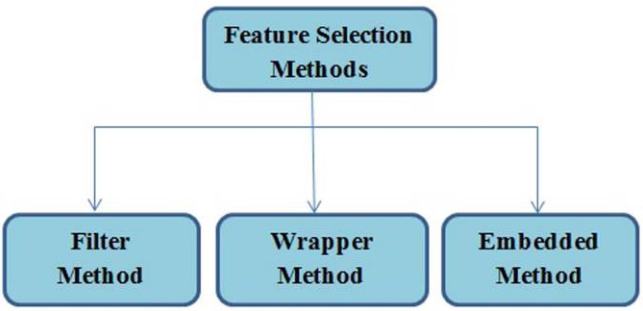}
	\caption{Feature Selection methods.}
	\label{fig:fsMethods}	
\end{figure}
Feature Selection methods can be grouped into two broad categories: classifier-independent (\textit{filters}) and classifier-dependent (\textit{wrappers} and \textit{embedded} methods) (see Fig. \ref{fig:fsMethods}). 
\begin{description}
	\item[Filters]  are based upon classifier-independent metrics such as distance, correlation, mutual information and consistency, which are used to rank each feature and remove the least ones. These methods are effective in term of time computation and give robustness to the overfitting phenomenon \cite{Hamon2013}.
	\item[Wrappers]  search the space of feature subsets, using the accuracy of a particular classifier as the measure of feasibility of a candidate subset. Such an approach allows to detect the possible interactions between variables \cite{Phuong2005} and select only the most relevant, non-redundant subset. However, it also has the disadvantage of a considerable computational cost, and may produce subsets that are overly specific to a specific classifier \cite{Brown2012Gavin}. As a result, any change in the learning model is likely to make the feature set sub optimal.

	\item[Embedded] methods exploit the structure of specific classes of learning models to guide the feature selection process \cite{Guyon2006}. In particular, the most common embedded methods make use of decision tree algorithms \cite{LetteriPC19}. Roughly speaking, in a decision tree, each internal node is associated with a feature, and the values of such feature are used to split the dataset in smaller subsets in the node's children. Actually, building a good decision tree implies making a good feature selection, since a good decision tree puts the most relevant features near to the root. Thanks to this, a decision three based learning algorithm can perform feature selection and classification simultaneously \cite{Saghapour2017}. Therefore, even if these methods are less computationally expensive and less prone to overfitting than wrappers, they are still tightly coupled with a specific learning model.
\end{description}

To sum up, filters are faster than embedded methods, and embedded methods are in turn faster than wrappers. In terms of overfitting, wrappers have higher learning capacity so they are more likely to overfit than embedded methods, which in turn are more likely to overfit than filter methods \cite{Brown2012Gavin}. Furthermore, the defining component of an embedded method is a criterion derived from the deep knowledge of a specific classification function, while the defining component of a wrapper method is simply the search procedure. In contrast, filter methods define a heuristic ranking criterion to act as a proxy measure of the classification accuracy and are independent from any particular classifier, thereby the selected features will be more generic, having incorporated less assumptions.

Filter methods can be further divided into two classes: univariate-based and multivariate-based \cite{Saeys2007Yvan}. The former have attracted much attention because of their low complexity and fast performance for high dimensionality. However, it is also known that some features discarded by univariate methods may be valuable for classification \cite{Guyon2003Elisseeff}, since such methods do not take into consideration the effects of feature-feature interactions. On the other hand, most of the current multivariate methods are bivariate-based filters which are almost based on entropy (or conditional entropy) and mutual information \cite{5432207}.

In this work we will develop a multivariate feature ranking methodology \cite {WangLijun2013}, by enhancing the filtering procedure shown in \cite {itasec2020}, where the Pearson correlation coefficient is used to measure feature dependency. Here, we exploit the mutual information as a more general correlation coefficient.

\subsection{Mutual Information Algorithms}


The mutual information algorithms employed in our FS strategy are all well-known. Given a set of features $X$ and a classification label $c$, they extract a subset $S \subseteq X$ as briefly recalled in the following.

\begin{enumerate}
    \item \textit{minimum Redundancy Maximum Relevance} (mRMR) \cite{1453511} tries to find the $S \subseteq X$ that has the highest relevance with respect to the target classification class $c$ while, at the same time, trying to minimize the information redundancy. More formally, the mRMR algorithm tries to find the optimal $S$ to maximize the relevance measure $\frac{1}{|S|}\sum_{S_i \in S}{I(S_i,c)}$ and minimize the redundancy measure $\frac{1}{|S|^2}\sum_{S_i,S_j \in S}{I(S_i,S_j)}$, where $I$ represents the mutual information between two features. 
    
    \item \textit{Joint Mutual Information} (JMI) \cite{BENNASAR2015} is based on conditional MI and selects features by checking whether they bring additional information to an existing feature set. This criterion considers second-order interactions between the features and the target classification $c$, allowing the detection of features which, when taken in pairs, provide more information about the output than the single features. In other words, JMI maximizes the amount $I(X,c) - \sum_{X_i \in S} (\alpha {I(X,X_i)} - \beta{I(X,X_i|c)}) $, where $\alpha=\beta=\frac{1}{|S|}$.
    
    JMI comes from the maximization as follows: $$X^{JMI} = arg max_{X_i \in X_{-S}} \{ u_i - z_i + c_i \}$$
    
    
    \item \textit{Double Input Symmetrical Relevance} (DISR) method is an updated JMI which adopts the so called double input symmetrical relevance criterion. It has shown to be one of the most effective MI-based feature selection criterion as  \cite{Brown2012Gavin}. DISR method select features finding the subset $S \subset {1, ..., n}$ that maximizes the mutual information $I(X_S,Y)$ in a greedy manner, the mutual entropy $H(X,Y)$, and defines the  \textit{symmetrical relevance} (SR) \cite{Meyer2006Bontempi} where $SR(X_S,Y)=\frac{I(X_S,Y)}{H(X_S,Y)}$, that indicates the concentration of mutual information between $X$ and $Y$ as follows:
    
    $$X_S^{DISR} = arg max_{X_S \in X} \{ \sum_{Xi \in X_S} \sum_{X_j \in X_S} SR(X_i,j,Y) \}$$
    
    \item \textit{Mutual Information-based Feature Selection} (MIFS) \cite{Battiti1994} is an iterative algorithm that finds the feature $X_i \in X$ that maximizes $I(X_i,c)$ and adds it to $S$. This process is repeated until $I(X_i,c)$ lowers beyond a given threshold. The selection is tuned by a proportional term $\beta I(X_i,S)$ that measures the information overlap between the candidate feature and existing features.\\
    
    
    \item \textit{Conditional Mutual Information Maximization} (CMIM) \cite{Fleuret2004,Wang2004} criterion harness process select the feature $X_i \in X_{-S}$ whose minimal relevance $I(X_i,Y|X_j)$ conditioned to the selected features $X_j \in X_S$, is maximal. Then, the minimal value is retained and the feature that has a maximal minimal conditional relevance is selected as follows:
    $$
        X_i^{CMIM}= arg max_{X_i \in X_{-S}} \{ min_{X_j \in X_S} I(X_i,Y|X_j) \}
    $$
    to greedily select dominant features.\\

    
    \item \textit{Conditional Information Feature Extraction} (CIFE) \cite{Akadi2008APF} defines the class-relevant redundancy $R_c(X_i,X_j)= I(X_i,X_j) - I(X_i,X_j|c)$ as the information carried a couple of features $X_i$ and $X_j$.  CIFE is calculated as follows: $ I(X_i,c) - \sum_{X_j \in S} R_c(X_i,X_j)$.
\end{enumerate}

\subsection{The Feature Selection Process}
The FS process we developed consists of the two steps described in the following.

\subsubsection{Dataset tampering.}

First, we add to the dataset three new random variables, with different distributions and independent from the target variable. Then, we use the six MI algorithms to rank the dataset variables in order of relevance. If any of these algorithms gives to one of the random variables an high ranking (with respect to a suitable threshold), it is removed from our algorithm suite.

\begin{figure*}[!ht] 
    \centering
    \subfigure[]{%
        \includegraphics[width=0.31\hsize]{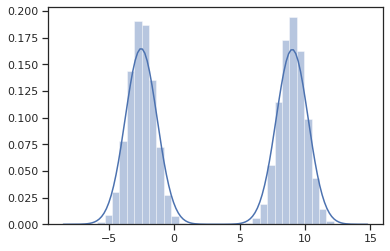}%
        \label{fig:RandFeat1}}%
    \quad%
    \subfigure[]{%
        \includegraphics[width=0.31\hsize]{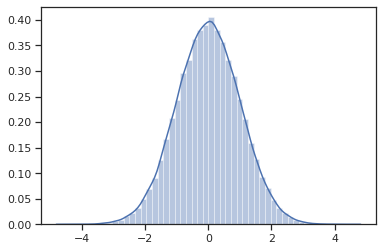}%
        \label{fig:RandFeat2}}%
    \quad%
    \subfigure[]{%
        \includegraphics[width=0.31\hsize]{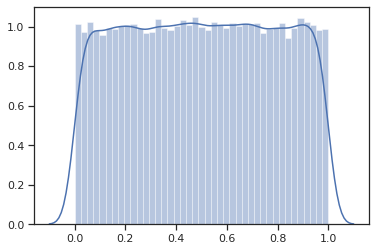}%
        \label{fig:RandFeat3}}%
    \caption{Value distribution of the three random features.}
    \label{fig:RandFeat}
\end{figure*}

In particular, we generate three random features as follows: 
\begin{itemize}
	\item The first feature is generated using the \textit{make\_gaussian\_quantiles} function \cite{mgq} (with the distribution shown in Fig. \ref{fig:RandFeat1}). 

	\item The second feature is generated using the \textit{make\_blobs} function \cite{mb} 
	(Fig. \ref{fig:RandFeat2}).

	\item The third feature is generated using the numpy \textit{uniform} library function \cite{ulf} 
	(Fig. \ref{fig:RandFeat3}). 
 \end{itemize}
 



The distributions  of such random variables are very different from the ones of the real MTA-KDD'19 dataset,
so we expect them to be easily identified as \textit{useless} for the traffic classification.


\begin{table}[ht]
\center 
\caption{Random feature rankings.}
\begin{tabular}{|l|l|l|l|}
\hline
                \textbf{Algorithm}           & \textbf{1st RandFeat} & \textbf{2nd RandFeat} & \textbf{3rd RandFeat} \\ \hline
\multicolumn{1}{|l|}{mRMR} &  $29^{th}$   &  $35^{th}$  &  $36^{th}$  \\ \hline
\multicolumn{1}{|l|}{MIFS} &  $21^{st}$   &  $35^{th}$  &  $36^{th}$  \\ \hline
\multicolumn{1}{|l|}{CIFE} &  $22^{nd}$   &  $33^{rd}$  &  $34^{th}$    \\ \hline  \hline

\multicolumn{1}{|l|}{JMI}  &  $19^{th}$   &  $2^{nd}$  &  $3^{rd}$   \\ \hline
\multicolumn{1}{|l|}{CMIM} &  $25^{th}$   &  $8^{th}$  &  $9^{th}$    \\ \hline
\multicolumn{1}{|l|}{DISR} &  $6^{th}$   &  $1^{st}$  &  $2^{nd}$    \\ \hline
\end{tabular}
\label{tab:ranks}
\end{table}

To make our experimentation more robust, we apply a 5-fold  cross validation, so we split our extended dataset in five parts, run the mutual information FS algorithms suite on each part and generate the final ranking by averaging the ranks obtained in the five experiments. In Tab. \ref{tab:ranks} we show the results relative to the three random variables only. The mRMR and MIFS algorithms responded well to the tampering, giving to such random features a low ranking in all the five experiments. Then we have the CIFE algorithm, which in some experiments gives to the random variables an higher ranking. We discard the other algorithms, which performed worse than CIFE.

\subsubsection{Backward Feature Elimination.} 

Back to the original dataset, for each algorithm left in our suite after the first step,  i.e., mRMR, MIFS and CIFE, we perform the following:


\begin{enumerate}
\item generate the feature ranking on the current dataset features
\item remove the lowest-ranked feature from the dataset 
\item evaluate 
the accuracy, precision and recall metrics obtained using a linear SVM on the resulting reduced dataset
\item if at least one of the three metrics goes below the threshold $\gamma$, stop the process, otherwise repeat from step 1.
\end{enumerate}
Note that we considered these metrics as quality/stopping criteria as they give valuable information also on unbalanced datasets as the MTA-KDD'19 \cite{MTAKDD19} that, being constantly fed with up-to-date malware traces, is likely to contain more malware than regular traffic  (at the time of our experiments, the dataset was however only slightly unbalanced: 46\% vs. 54\%). In our experiments, we set  $\gamma=97\%$ which is a reasonable threshold given the number of samples and the specific ML model adopted to calculate the metrics.  

\begin{figure}[ht]
\centering
{\includegraphics[width=0.8\hsize]{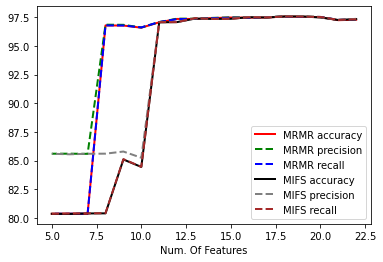}} \quad
\caption{SVM metrics during backward feature elimination with the mRMR and MIFS rankings.}
\label{fig:svmMean}
\end{figure}

Fig. \ref{fig:svmMean} shows the accuracy, precision and recall metrics calculated during the elimination process from 23 to 5 features, using the mRMR and MIFS rankings. All the performance indicators remain stable above $97\%$ for subsets composed by 23 or more features, even if  MIFS appears slightly less accurate than mRMR. From 23 to 11 features the metrics start lowering slowly, and finally they quickly move below $97\%$ when the dataset is reduced to less than 11 features. This quick decay is visible also in the CIFS metrics (not shown here). Therefore, we may conclude that 11 is the minimum number of features that allows us to preserve a reasonable classification accuracy. In the remaining part of this paper, we will refer to this threshold as the \textit{Maximum Dimension Restriction threshold} (MDRt). 

\begin{table}[!ht]
\centering  
\caption{Accuracy, Precision, and Recall measured after reducing the detaset to 11 features with each MI algorithm ranking.}
\begin{tabular}{|c|c|c|c|}
\hline
\textbf{Algorithm}              & \textbf{Accuracy} & \textbf{Precision} & \textbf{Recall} \\ \hline
mRMR & 97.07                             & 97.09                              & 97.07                           \\ \hline
MIFS & 97.06                             & 97.08                              & 97.06      \\ \hline \hline

CIFE & 88.46                             & 88.70                              & 86.41      \\  \hline \hline

JMI  & 84.35                             & 85.25                              & 84.35                           \\ \hline
CMIM & 88.19                             & 88.52                              & 88.21                           \\ \hline
DISR & 81.27                             & 81.56                              & 81.26                           \\ \hline

\end{tabular}
 \label{tab:AvgAccPr}
 \end{table}
 
In particular, Tab.  \ref{tab:AvgAccPr} reports the metrics measured after reducing the features to the MDRt using each MI algorithm ranking. Actually, for sake of completeness, we performed the backward elimination experiment with all the six MI algorithms. The results show that the three algorithms removed in the previous step would have indeed  very bad results in this phase, and that also the CIFE results in significantly lower metrics than MIFS and mRMR. Therefore, we removed also the CIFE from out algorithms suite.

\begin{table}[!ht]
\center  
\caption{Ranking of mRMR and MIFS on the 11 most relevant features.}
\begin{tabular}{l|l|l|l|l|l|l|}
\cline{2-5}
     & \multicolumn{2}{c|}{\textbf{mRMR}}   & \multicolumn{2}{c|}{\textbf{MIFS}} \\ 
     \cline{2-5} 
     & \multicolumn{1}{c|}{\textbf{Feature}} & \multicolumn{1}{c|}{\textbf{Score}} & \multicolumn{1}{c|}{\textbf{Feature}} & \multicolumn{1}{c|}{\textbf{Score}} \\ \hline
\multicolumn{1}{|l|}{\textit{1st}}  & StartFlow  & 0.3710   & StartFlow  & 0.3516   \\ \hline
\multicolumn{1}{|l|}{\textit{2nd}}  & NumIPdst  & 0.3417    & NumIPdst  & 0.3193    \\ \hline
\multicolumn{1}{|l|}{\textit{3rd}}  & NumCon  & 0.3392   & NumCon   & 0.3218    \\ \hline
\multicolumn{1}{|l|}{\textit{4th}}  & NumPorts  & 0.3101   & NumPorts  & 0.2909   \\ \hline
\multicolumn{1}{|l|}{\textit{5th}}  & MinLenrx  & 0.2931   & FirstPktLen  & 0.2179  \\ \hline
\multicolumn{1}{|l|}{\textit{6th}}  & FirstPktLen  & 0.2516   & MinLen  & 0.2134   \\ \hline
\multicolumn{1}{|l|}{\textit{7th}}  & MinLen  & 0.1327   & TCPoverIP  & 0.0983  \\ \hline
\multicolumn{1}{|l|}{\textit{8th}}  & UDPoverIP  & 0.1253   & MinLenrx   & 0.0857   \\ \hline
\multicolumn{1}{|l|}{\textit{9th}}  & TCPoverIP & 0.0972   & DNSoverIP  & 0.0738   \\ \hline
\multicolumn{1}{|l|}{\textit{10th}} & DNSoverIP  & 0.0895   & RstFlagDist  & 0.0615  \\ \hline
\multicolumn{1}{|l|}{\textit{11th}} & RstFlagDist  & 0.0687   & UDPoverIP   & 0.0423    \\ \hline
\end{tabular}
\label{tab:2algosRank}
\end{table}

Looking in detail at the top-11 feature rankings given by  mRMR and MIFS and reported in Tab. \ref{tab:2algosRank}, we can see that they contain the same features. In particular, the top-four features are exactly the same, whereas the remaining seven are the same but in a slightly different order.  Therefore, we extract dataset composed by the above $MDRt$ features, calling it the \textit{Optimized MTA-KDD'19 dataset}.

\subsection{Rank-Relevance Weight and Sensibility Enhancement}

To further refine the Optimized MTA-KDD'19 dataset, we assign each feature $f$ a \textit{Rank Relevance-weighted} (RRw) score derived as follows:

$$
RRw(f) = \left\Vert \frac{\sum_{i=1}^{n} s_i(f) \cdot avgF1^k(i)}{n} \right\Vert
$$

\noindent where $s_i(f)$ is the score given to $f$ by algorithm $i$ (in our case we have $n=2$ algorithms), whereas the $avgF1^k_i$ is the average F1-score (i.e., the weighted harmonic mean of precision and recall) derived by the $k=5$ cross folder validation on the 11-features dataset obtained with the $i$-th algorithm ranking. The final results are normalized in the range (0,1] using a \textit{MinMax} normalization.

We scale the sample values by multiplying each by the corresponding feature RRw score, calling the resulting dataset \textit{RRw-Optimized MTA-KDD'19 dataset}.

Such RRw-based scaling acts as a \textit{sensibility enhancing criterion}. Indeed, the most important features are scaled up (having a higher RRw), making them more relevant to the classifier. From the optimizer point of view, the gradient will be higher in the directions given by the upscaled features, allowing it to more quickly reach the minimum classification loss. More formally, in general the sensibility $R_i$ of a model $f(x)$ relative to a feature $x_i$ can be written as:
$$
  R_i = \left(\frac{ \partial f} { \partial x_i }\right)^2  
$$


using the chain rule, we can show that for a function which is locally differentiable the sensibility for the new transformed dataset will be:

$$
R_i^w = \left(\frac{\partial f( w_1 x_1, ..., w_i x_i, ... w_N x_N)}{ \partial x_i}\right)^2 = \left(w_i  \frac{ \partial f}{ \partial x_i}\right)^2 =  w_i^2  R_i
$$
Therefore, weighting a feature \textit{enhances the model sensibility to such feature quadratically to its assigned weight}. 


\section{Dimensional Reduction through Autoencoder}

Our second dataset optimization approach makes use of the ability of AE to encode data into lower-dimensional codes. Indeed, AE are optimized to learn, during their training, an efficient, compressed representation of the input data (obtained by nonlinearly combining such data with different weights) in their internal \textit{bottleneck layer}. 

Then, if we minimize the reconstruction error (i.e., the error introduced when decompressing data from the bottleneck layer to the output layer) and the group sparsity regularization \cite{Yuan2006Ming} simultaneously, especially in neural networks models \cite{SCARDAPANE201781}, we can extract from, the bottleneck layer a new, smaller set of features which  preserves the intrinsic information of the original data. 

To this aim, AE has 33 input and output neurons (as the total number of features) and, for the bottleneck layer, we choose a size equal to the MDRt derived with the FS methodology (11), which seems to be a reasonable compression threshold. 

The current MTA-KDD'19 dataset contains 64554 samples. We split it in an 85\% training set (54870 samples) and a 15\% \textit{validation set} (9684 samples). Moreover, the training set is further split into a 15\% (8231 samples) \textit{testing set} and a 85\% \textit{learning set} (46639 samples). The AE training will be performed using the learning and validation sets, whereas the testing set (which is never fed to the network during training) will be used for the final performance tests.

\begin{figure}[ht]
  \centering
  \includegraphics[width=0.8\hsize]{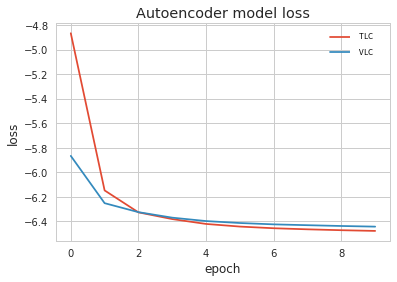}
  \caption{Autoencoder reconstruction error.}
  \label{fig:AElosses}
\end{figure}

The AE is trained for 10 epochs as done in \cite{itasec2020} but, from Fig. \ref{fig:AElosses}, it is possible to see that the model loss (i.e., the data decompression/reconstruction error) converges nicely and the error becomes acceptable from the fourth epoch.

\subsection{Latent AE MTA-KDD'19 features}
Once the training is completed with satisfying results, use the AE as a "feature compressor". To this aim, we feed it with a 33-feature sample and read its corresponding 11-dimensional latent representation from the bottleneck layer.


\begin{figure*}[!ht] 
    \centering
    \subfigure[RRw OptMTA-KDD Pair Plot]{%
        \includegraphics[width=0.49\hsize]{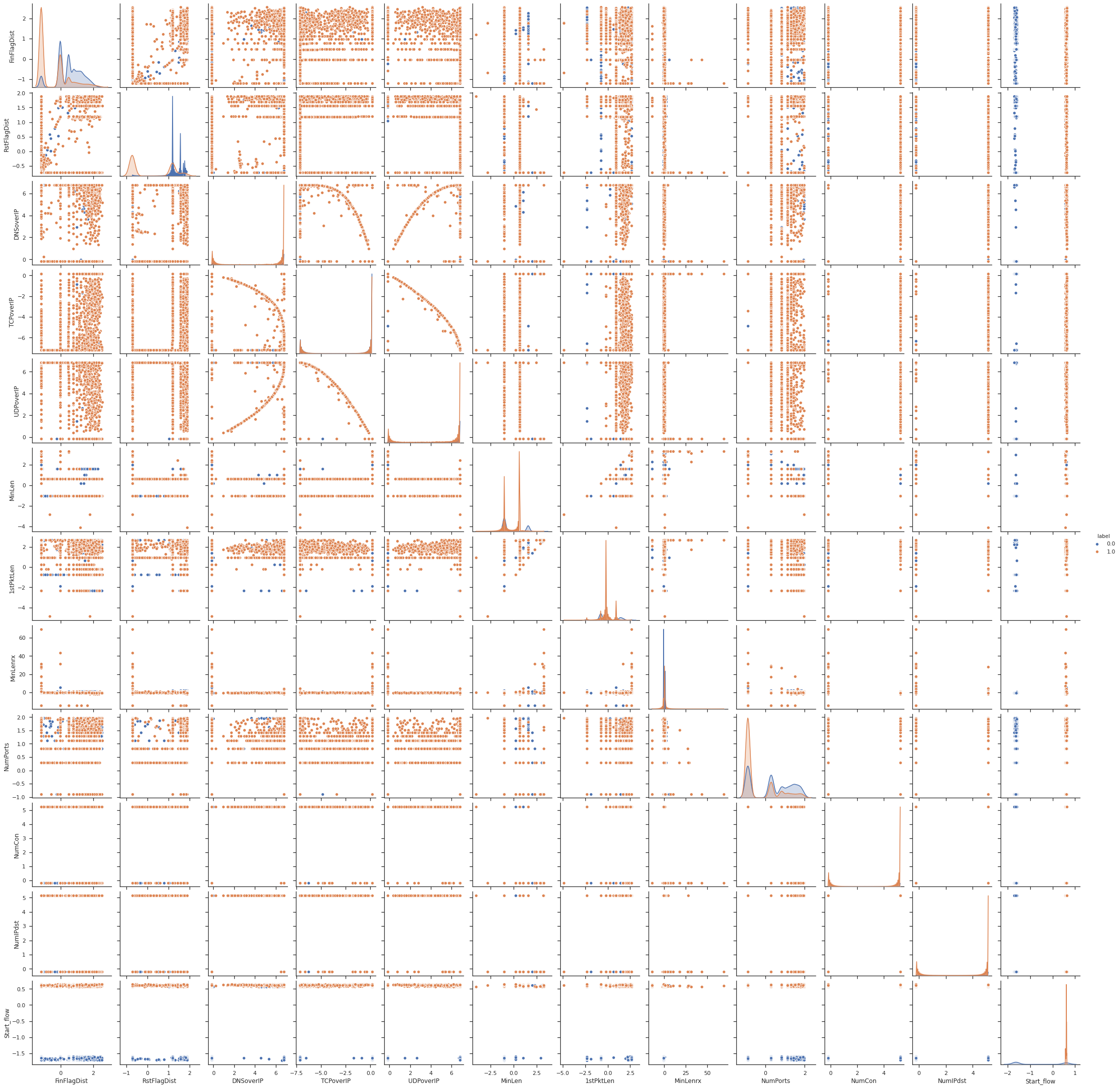}%
        \label{fig:ppRPwOptMTAKDD19}}%
    \quad%
    \subfigure[AE MTA-KDD Pair Plot]{%
        \includegraphics[width=0.47\hsize]{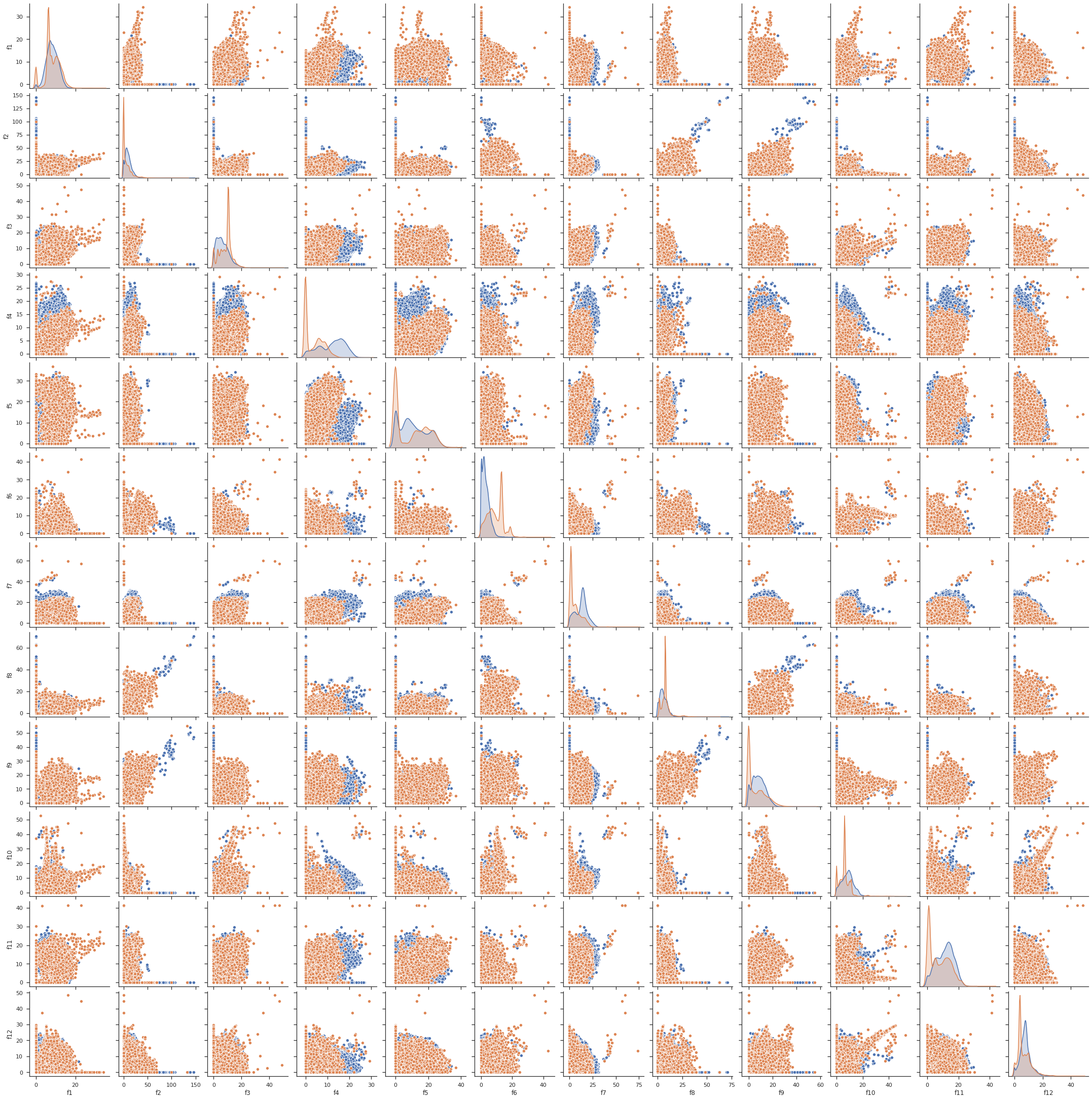}%
        \label{fig:PPMdsA}}%
    \caption{Comparison between the RRw Optimized MTA-KDD dataset and AE MTA-KDD'19 dataset.}
    \label{fig:PairPlotsADSM}
\end{figure*}

Fig. \ref{fig:ppRPwOptMTAKDD19} shows the scatter plot matrix of the RRw-Optimized MTA-KDD'19 features composed by 11 features, from bottom to top, in the following sequence: \textit{NumIPdst}, \textit{NumCon}, \textit{NumPorts}, \textit{MinLenrx}, \textit{1stPktLen}, \textit{MinLen}, \textit{UDPoverIP}, \textit{TCPoverIP}, \textit{DNSoverIP}, \textit{RstFlagDist} and \textit{FinFlagDist}. All details about these features are discussed in \cite{itasec2020}.

The saliency of features is determined by their relevance, redundancy, and density distributions.
Fig. \ref{fig:PairPlotsADSM} shows the relation between the features involved and allow to see on the diagonals the density distributions of the samples. In the plots, blue dots represent legitimate traffic samples whereas red dots represent malware samples.

Fig. \ref{fig:PPMdsA} shows the pair plot of the latent AE MTA-KDD'19 features. By comparing this plot with the one in Fig. \ref{fig:ppRPwOptMTAKDD19}, it is possible to see that the new set of features is actually different from the original one where the separation between classes is particularly clear especially in three features. The AE mixed the distribution of the samples and changed the range of values, creating only sporadic small clusters and less outliers with respect to the RRw-Optimized MTA-KDD'19 dataset. Moreover, it is worth noting that the spikes relative to the malware traffic have been preserved, whereas legitimate traffic has a Gaussian-like distribution skewed to the left.

\begin{figure}[!ht]
	\centering
	\includegraphics[width=0.8\hsize]{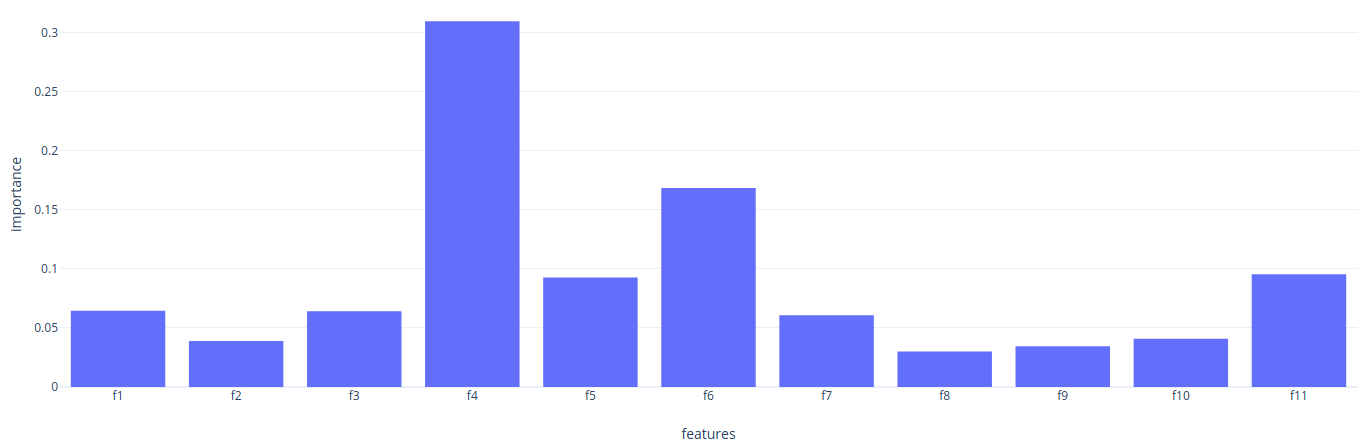}
	\caption{Average of Feature Importance with four Decision Trees.}
	\label{fig:featimp}
\end{figure}

It is worth noting that, in the AE MTA-KDD'19 dataset, all the features are now relevant, as shown in Fig. \ref{fig:featimp}. 
Note that, on this kind of dataset, it is now very hard to apply further optimizations (if any), as the connections between the generated features and the original ones is lost, for this reason it is reported in the fig. \ref{fig:PPMdsA} with anonymous label from $f1$ to $f11$.

\section{Experimentation}
In this section we show the experiments performed to measure the accuracy of the Optimized, RRw-Optimized, and AE-generated MTA-KDD'19 datasets. The experiments performed in \cite{itasec2020} on the full 33-features MTA-KDD'19 dataset will be used as our baseline.
Therefore, we reuse the same MLP described in \cite{itasec2020}, i.e., a rectangle-shaped fully connected MLP with two hidden layers, both with $2f$ neurons (where $f$ which is the number of features in the dataset), and a single-neuron output layer with sigmoid activation function. Here, obviously, we set $f=11$ in order to shrink the MLP according to the new dataset features size. The MLP is trained for 10 epochs with batch size equal to 10. 

The datasets to evaluate are split following the same criteria introduced in the AE experiments, that is, 85\% train set, further split in a 15\% (8231 samples) test set and a 85\% learning set (46639 samples), and a 15\% validation set (9684 samples). 

To evaluate the model performances with respect to a specific dataset, we consider its Train Learning Curve (TLC, showing the loss evolution during the training phase of each epoch, i.e., of how well the model is learning) and its Validation Learning Curve (VLC, showing the loss evolution during the validation phase at the end of each epoch, i.e., how well the model is generalizing).

\begin{figure}[!ht]
	\centering
	{\includegraphics[width=\textwidth]{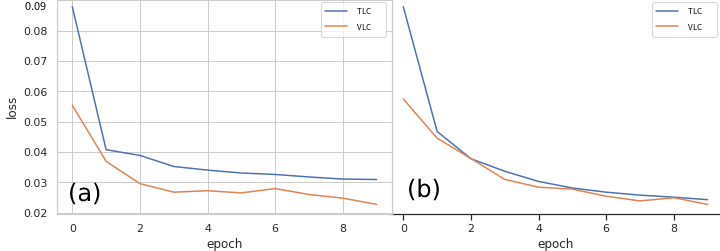}} 
	\caption{TLC and VLC on the Optimized (a) and  RRw-Optimized (b) MTA-KDD'19 datasets.}
	\label{fig:dnnRPw}
\end{figure}

The left side (a) of Fig. \ref{fig:dnnRPw} shows the learning curves of the MLP trained with the Optimized and RRw-Optimized MTA-KDD'19 datasets. We can see that the model underfits: the loss continues to decrease at the end of the plot, indicating that the model is capable of further learning and therefore that the training process was halted prematurely. With further experiments (not shown here), we determined that the loss stabilizes after 100 epochs.

The right side (b) of Fig. \ref{fig:dnnRPw} shows the learning curves of the MLP trained with the RRw-Optimized MTA-KDD'19 dataset. Here the TLC and VLC evolve similarly and the neural network loss stabilizes earlier at an acceptable level. Thus, as expected, re-weighting the dataset helped the classifier. 

\begin{figure}[!ht]
    \centering
    \includegraphics[width=\textwidth]{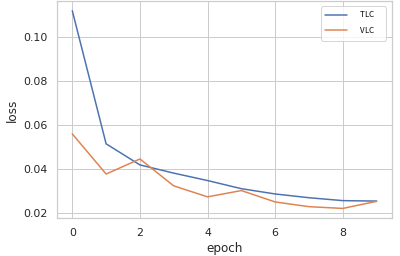} 
   \caption{TLC and VLC on the AE-Generated MTA-KDD'19 dataset.}
    \label{fig:AEtraintest}
\end{figure}

On the other hand, Fig. \ref{fig:AEtraintest} shows that with the AE-Generated dataset the error reduction is slower than with the RRw-Optimized one, but at the tenth epoch the loss is almost stable.

\begin{table}[ht]
\centering 
\caption{Final metrics on the testing set with the Optimized, RRw-Optimized and AE-Generated MTA-KDD'19 datasets.}
\begin{tabular}{|l|c|c|c|c|c|c|r|}
\hline
 & \textbf{Precision} & \textbf{TPR} & \textbf{TNR} & \textbf{FPR} & \textbf{FNR} & \textbf{FDR} & \textbf{Accuracy} \\ \hline
\textit{Optimized} & 99.40            & 99.48        & 99.31        & 0.069         & 0.052        & 0.060         & 99.40             \\ \hline
\textit{RRw-Optimized} & 99.63           & 99.61        & 99.59        & 0.041         & 0.039        & 0.037         & 99.60             \\ \hline

\textit{AE-Generated} & 98.78       & 99.61        & 98.60        & 0.14         & 0.039         & 0.12         & 99.14             \\ \hline
\end{tabular}
\label{tab:MLPmetrics}
\end{table}



Tab. \ref{tab:MLPmetrics} shows the performance metrics measured after the tenth epoch on the testing set with the three datasets. In particular, the \textit{True Positive Rate} (TPR)/\textit{True Negative Rate} (TNR), and the \textit{False Positive Rate} (FPR)/\textit{False Negative Rate} (FNR)  measure the proportion of  positives/negatives that are correctly/wrongly identified, respectively. The \textit{False Discovery Rate} (FDR) indicates the rejection rate of the false positives \cite{Colquhoun2014}.

We can see that the RRw-Optimization improves \textit{all} the metrics. Furthermore, with the RRw-Optimized dataset, the \textit{generalization gap} (i.e., the distance between training and validation loss) is minimal. The AE-Generated dataset results in a triple false positive rate, but its the overall performance metrics, even if slightly lower, remain acceptable.

\section{Conclusions}
In the context of machine learning-assisted malware traffic analysis, selecting a small set of meaningful features to train the classifier is a crucial task \cite{Letteri2020Journal}. Indeed, the very variable results presented by most of the literature works in this field may be also due to the use of different combinations of such features.

On the other hand, several feature selection and dimensional reduction algorithms have been presented, and some of them could be profitably applied in the optimization of the network traffic datasets.

In this paper we present two optimization strategies specifically tailored to (malware) traffic datasets, and in particular to the newly presented MTA-KDD'19 dataset \cite{MTAKDD19}, which exploit and combine the most promising state-of-the art feature selection and dimensional reduction approaches. 

The experiments show that the first, FS-based optimization approach exploiting MI is expensive when calculating the feature relevance and the MDRt, but the corresponding \textit{RRw-optimized} dataset achieves the highest accuracy. On the other hand, the second, AE-based optimization reveals a good compromise between pre-processing time and accuracy of the derived \textit{AE-generated} dataset.

Therefore, both approaches may be a good choice in specific usage scenarios, possibly integrated in an anti-virus (AV) or in combination with an intrusion detection system (IDS). An offline-trained classifier may profitably exploit the RRw-optimized dataset because it requires more computational time in the pre-processing step but it is more efficient with its 99,60\% of accuracy. On the other hand, advanced approaches such as using a MLP as an online-training for a sort of \textit{near real-time} detection, and \textit{dynamic-training} AV and/or IDS, as proposed in the MTA-KDD'19 project \cite{itasec2020}, requires the model to be \textit{small} and \textit{fast}. Indeed, the dynamic training requires the model to receive, at regular intervals, new input samples to train with and thus update its classification capabilities. Such periodic training must thus be as quick as possible and at the same time preserves the highest possible accuracy like could be 99,14\% achieved from our AE, so the AE-Generated dataset would be a better choice with the right compromise between accuracy detection and time training.\\

Both the RRw-Optimized and AE-Generated MTA-KDD’19 datasets will be made publicly available as a fork of GitHub repository \cite{RRw-OptMTAKDD19} of the authors. As a future work, we plan to apply more complex balancing techniques to these datasets in order to further improve the classification accuracy.

\bibliographystyle{splncs04}
\bibliography{main}

\begin{thebibliography}{10}
\providecommand{\url}[1]{\texttt{#1}}
\providecommand{\urlprefix}{URL }
\providecommand{\doi}[1]{https://doi.org/#1}

\bibitem{Akadi2008APF}
Akadi, A.E., Ouardighi, A.E., Aboutajdine, D.: A powerful feature selection
  approach based on mutual information (2008)

\bibitem{RRw-OptMTAKDD19}
Authors: {Optimized MTA-KDD'19} datasets (2020), {URL hidden for paper
  anonymization purposes}

\bibitem{5432207}
{Balagani}, K.S., {Phoha}, V.V.: On the feature selection criterion based on an
  approximation of multidimensional mutual information. IEEE Transactions on
  Pattern Analysis and Machine Intelligence  \textbf{32}(7),  1342--1343 (2010)

\bibitem{Battiti1994}
Battiti, R.: Using mutual information for selecting features in supervised
  neural net learning. Neural Networks, IEEE Transactions on  \textbf{5},  537
  -- 550 (08 1994). \doi{10.1109/72.298224}

\bibitem{BENNASAR2015}
Bennasar, M., Hicks, Y., Setchi, R.: Feature selection using joint mutual
  information maximisation. Expert Systems with Applications  \textbf{42}(22),
  8520 -- 8532 (2015). \doi{10.1016/j.eswa.2015.07.007},
  \url{http://www.sciencedirect.com/science/article/pii/S0957417415004674}

\bibitem{Borges2012}
Borges, H.B., Nievola, J.C.: Comparing the dimensionality reduction methods in
  gene expression databases. Expert Syst. Appl.  \textbf{39}(12),
  10780–10795 (Sep 2012). \doi{10.1016/j.eswa.2012.03.015},
  \url{https://doi.org/10.1016/j.eswa.2012.03.015}

\bibitem{Brown2012Gavin}
Brown, G., Pocock, A., Zhao, M.J., Luján, M.: Conditional likelihood
  maximisation: A unifying framework for information theoretic feature
  selection. The Journal of Machine Learning Research  \textbf{13},  27--66 (02
  2012)

\bibitem{Colquhoun2014}
Colquhoun, D.: An investigation of the false discovery rate and the
  misinterpretation of p-values. Royal Society Open Science  \textbf{1}(3),
  140216 (Nov 2014). \doi{10.1098/rsos.140216},
  \url{http://dx.doi.org/10.1098/rsos.140216}

\bibitem{Fleuret2004}
Fleuret, F.: Fast binary feature selection with conditional mutual information.
  Journal of Machine Learning Research  \textbf{5},  1531--1555 (11 2004)

\bibitem{Gareth2013}
Gareth~James, Daniela~Witten, T.H., Tibshirani, R.: An Introduction to
  Statistical Learning, p.~204. Springer Texts in Statistics 103 (2013).
  \doi{10.1007/978-1-4614-7138-7}

\bibitem{Guyon2003Elisseeff}
Guyon, I., Elisseeff, A.: An introduction to variable and feature selection. J.
  Mach. Learn. Res.  \textbf{3}(null),  1157–1182 (Mar 2003)

\bibitem{Guyon2006}
Guyon, I., Gunn, S., Nikravesh, M., Zadeh, L.: Feature extraction: foundations
  and applications (01 2006)

\bibitem{Hamon2013}
Hamon, J.: Optimisation combinatoire pour la s{\'e}lection de variables en
  r{\'e}gression en grande dimension : Application en g{\'e}n{\'e}tique
  animale. (combinatorial optimization for variable selection in high
  dimensional regression: Application in animal genetic) (2013)

\bibitem{Han2017AutoencoderFS}
Han, K., Li, C., Shi, X.: Autoencoder feature selector. ArXiv
  \textbf{abs/1710.08310} (2017)

\bibitem{Han2018}
Han, K., Wang, Y., Zhang, C., Li, C., Xu, C.: Autoencoder inspired unsupervised
  feature selection. 2018 IEEE International Conference on Acoustics, Speech
  and Signal Processing (ICASSP) pp. 2941--2945 (2018)

\bibitem{1453511}
{Hanchuan Peng}, {Fuhui Long}, {Ding}, C.: Feature selection based on mutual
  information criteria of max-dependency, max-relevance, and min-redundancy.
  IEEE Transactions on Pattern Analysis and Machine Intelligence
  \textbf{27}(8),  1226--1238 (2005)

\bibitem{HintonZemel1993}
Hinton, G.E., Zemel, R.S.: Autoencoders, minimum description length and
  helmholtz free energy. In: Proceedings of the 6th International Conference on
  Neural Information Processing Systems. p. 3–10. NIPS’93, Morgan Kaufmann
  Publishers Inc., San Francisco, CA, USA (1993)

\bibitem{6129431}
{Huang}, Y., {Xu}, D., {Nie}, F.: Semi-supervised dimension reduction using
  trace ratio criterion. IEEE Transactions on Neural Networks and Learning
  Systems  \textbf{23}(3),  519--526 (March 2012).
  \doi{10.1109/TNNLS.2011.2178037}

\bibitem{1054102}
{Hughes}, G.: On the mean accuracy of statistical pattern recognizers. IEEE
  Transactions on Information Theory  \textbf{14}(1),  55--63 (January 1968).
  \doi{10.1109/TIT.1968.1054102}

\bibitem{MTAKDD19}
Letteri, I.: {MTA-KDD'19} dataset (2019),
  \url{https://github.com/IvanLetteri/MTA-KDD-19}

\bibitem{LetteriPC19}
Letteri, I., {Della Penna}, G., Caianiello, P.: Feature selection strategies
  for {HTTP} botnet traffic detection. In: 2019 {IEEE} European Symposium on
  Security and Privacy Workshops, EuroS{\&}P Workshops 2019, Stockholm, Sweden,
  June 17-19, 2019. pp. 202--210. {IEEE} (2019).
  \doi{10.1109/EuroSPW.2019.00029}

\bibitem{Letteri2018CSS}
Letteri, I., Della~Penna, G., De~Gasperis, G.: Botnet detection in software
  defined networks by deep learning techniques. In: Castiglione, A., Pop, F.,
  Ficco, M., Palmieri, F. (eds.) Cyberspace Safety and Security. pp. 49--62.
  Springer International Publishing, Cham (2018).
  \doi{10.1007/978-3-030-01689-0\_4}

\bibitem{Letteri2020Journal}
Letteri, I., {Della Penna}, G., {De Gasperis}, G.: Security in the internet of
  things: botnet detection in software-defined networks by deep learning
  techniques. In: 2020 {IJHPCN} International Journal of High Performance
  Computing and Networking. vol.~15, pp. 170--182. {IJHPCN} (2020).
  \doi{10.1504/IJHPCN.2019.106095},
  \url{https://doi.org/10.1504/IJHPCN.2019.106095}

\bibitem{itasec2020}
Letteri, I., {Della Penna}, G., Vita, L.D., Grifa, M.T.: Mta-kdd'19: {A}
  dataset for malware traffic detection. In: Loreti, M., Spalazzi, L. (eds.)
  Proceedings of the Fourth Italian Conference on Cyber Security, Ancona,
  Italy, February 4th to 7th, 2020. {CEUR} Workshop Proceedings, vol.~2597, pp.
  153--165. CEUR-WS.org (2020), \url{http://ceur-ws.org/Vol-2597/paper-14.pdf}

\bibitem{Lu2018}
Lu, Q., Qiao, X.: Sparse fisher's linear discriminant analysis for partially
  labeled data. Statistical Analysis and Data Mining  \textbf{11},  17--31
  (2018)

\bibitem{Meyer2006Bontempi}
Meyer, P.E., Bontempi, G.: On the use of variable complementarity for feature
  selection in cancer classification. In: Proceedings of the 2006 International
  Conference on Applications of Evolutionary Computing. p. 91–102.
  EuroGP’06, Springer-Verlag, Berlin, Heidelberg (2006).
  \doi{10.1007/11732242\_9}

\bibitem{ulf}
Numpy: numpy.random.uniform. https://numpy.org/numpy.random.uniform.html

\bibitem{Pasunuri2020}
Pasunuri, R., Venkaiah, V.C.: A computationally efficient data-dependent
  projection for dimensionality reduction. In: Bansal, J.C., Gupta, M.K.,
  Sharma, H., Agarwal, B. (eds.) Communication and Intelligent Systems. pp.
  339--352. Springer Singapore, Singapore (2020)

\bibitem{Phuong2005}
Phuong, T.~M., L.Z..A.R.B.: Choosing snps using feature selection. Proceedings.
  IEEE Computational Systems Bioinformatics Conference p. 301–309 (2005).
  \doi{10.1109/csb.2005.22}

\bibitem{Saeys2007Yvan}
Saeys, Y., Inza, I., Larranaga, P.: A review of feature selection techniques in
  bioinformatics. Bioinformatics (Oxford, England)  \textbf{23},  2507--17 (11
  2007). \doi{10.1093/bioinformatics/btm344}

\bibitem{Saghapour2017}
Saghapour~E, Kermani~S, S.M.: A novel feature ranking method for prediction of
  cancer stages using proteomics data. PLoS One  (09 2017).
  \doi{10.1371/journal.pone.0184203}

\bibitem{SCARDAPANE201781}
Scardapane, S., Comminiello, D., Hussain, A., Uncini, A.: Group sparse
  regularization for deep neural networks. Neurocomputing  \textbf{241},  81 --
  89 (2017). \doi{10.1016/j.neucom.2017.02.029},
  \url{http://www.sciencedirect.com/science/article/pii/S0925231217302990}

\bibitem{mb}
Scikit-Learn: make\_blobs. https://scikit-learn.org
  /stable/modules/generated/sklearn.datasets.make\_blobs.html

\bibitem{mgq}
Scikit-Learn: make\_gaussian\_quantiles. https://scikit-learn.org
  /stable/modules/generated/sklearn.datasets.make\_gaussian\_quantiles.html

\bibitem{7530147}
{Shahana}, A.H., {Preeja}, V.: Survey on feature subset selection for high
  dimensional data. In: 2016 International Conference on Circuit, Power and
  Computing Technologies (ICCPCT). pp.~1--4 (2016)

\bibitem{Sorzano2014}
Sorzano, C.O.S., Vargas, J., Pascual-Montano, A.D.: A survey of dimensionality
  reduction techniques. ArXiv  \textbf{abs/1403.2877} (2014)

\bibitem{Wang2004}
Wang, G., Lochovsky, F.: Feature selection with conditional mutual information
  maximin in text categorization. pp. 342--349 (01 2004).
  \doi{10.1145/1031171.1031241}

\bibitem{WangLijun2013}
Wang, L., Lei, Y., Zeng, Y., Tong, l., Yan, B.: Principal feature analysis: A
  multivariate feature selection method for fmri data. Computational and
  mathematical methods in medicine  \textbf{2013},  645921 (09 2013).
  \doi{10.1155/2013/645921}

\bibitem{Wang2017Shuyang}
Wang, S., Ding, Z., Fu, Y.: Feature selection guided auto-encoder. In:
  Proceedings of the Thirty-First AAAI Conference on Artificial Intelligence.
  p. 2725–2731. AAAI’17, AAAI Press (2017)

\bibitem{Yuan2006Ming}
Yuan, M., Lin, Y.: Model selection and estimation in regression with grouped
  variables. Journal of the Royal Statistical Society Series B  \textbf{68},
  49--67 (02 2006). \doi{10.1111/j.1467-9868.2005.00532.x}

\end{thebibliography}
\end{document}